\documentclass[10pt, a4paper]{article}

\usepackage{lrec-coling2024} 

\usepackage{lingmacros}
\usepackage{amsmath,amssymb}
\usepackage{mathtools}
\usepackage{booktabs}
\usepackage{enumerate}
\usepackage{enumitem}
\usepackage{multirow}
\usepackage{subcaption}
\usepackage{floatrow}
\usepackage[whole]{bxcjkjatype}
\usepackage{expex}
\usepackage{url} 

\title{To Drop or Not to Drop? Predicting Argument Ellipsis Judgments: A Case Study in Japanese\\ \vspace*{.5\baselineskip}}

\name{Yukiko Ishizuki$^{1,2}$\quad Tatsuki Kuribayashi$^{3}$\quad Yuichiroh Matsubayashi$^{1,2}$ \quad \\
        \bf\large{Ryohei Sasano$^{4,2}$}\quad \bf\large{Kentaro Inui$^{3,1,2}$}}

\address{${}^{1}$Tohoku University, ${}^{2}$RIKEN, ${}^{3}$MBZUAI,  ${}^{4}$Nagoya University \\
         yukiko.ishizuki.p7@dc.tohoku.ac.jp, y.m@tohoku.ac.jp\\
         sasano@i.nagoya-u.ac.jp,  \{tatsuki.kuribayashi, kentaro.inui\}@mbzuai.ac.ae}

\abstract{
Speakers sometimes omit certain arguments of a predicate in a sentence; such omission is especially frequent in pro-drop languages.
This study addresses a question about ellipsis---what can explain the native speakers' ellipsis decisions?---motivated by the interest in human discourse processing and writing assistance for this choice.
To this end, we first collect large-scale human annotations of \textit{whether} and \textit{why} a particular argument should be omitted across over 2,000 data points in the balanced corpus of Japanese, a prototypical pro-drop language.
The data indicate that native speakers overall share common criteria for such judgments and further clarify their quantitative characteristics, e.g., the distribution of related linguistic factors in the balanced corpus.
Furthermore, the performance of the language model–based argument ellipsis judgment model is examined, and the gap between the systems' prediction and human judgments in specific linguistic aspects is revealed.
We hope our fundamental resource encourages further studies on natural human ellipsis judgment.
\\
\newline \Keywords{Argument Ellipsis, Language Resource, Discourse Processing, Japanese}}

\begin{document}

\maketitleabstract
\section{Introduction}
\label{sec:intro}
In pro-drop languages, arguments are often omitted from sentences, contrasting markedly with English where a subject is typically mandated~\citep{Van_Craenenbroeck2019-es}.
For example, our corpus analysis revealed that 37\% of arguments, such as subjects and objects in Japanese, are omitted (\S\ref{subsec:data}).
This frequent occurrence of \textit{argument ellipsis} or \textit{omission}\footnote{We use both terms interchangeably in this paper.}  has raised several challenges in natural language processing (NLP), such as recovering the omitted information to address the textual ambiguity~\cite{sasano-kurohashi-2011-discriminative,wang2018translating}.

In this study, we tackle a new question toward this phenomenon---\textit{when is such a context-dependent ellipsis preferred by native speakers and which linguistic factor is associated with this decision?}
The identification of human consensus on argument ellipsis judgments and its factors will contribute to linguistics and education: it can clarify the underlying consensus on the judgments among native speakers and be a helpful resource for writing assistance. 
Furthermore, it may be possible to explore applications such as an automatic text proofreading system can be explored based on the findings.

This paper presents the first large-scale study wherein the aforementioned question was investigated within a balanced corpus of Japanese.
Japanese is a prototypical pro-drop language, which is typically adopted to study the ellipsis-related phenomena~\cite{Iida2007-fx, Shibata2018-en, Konno2021-lu}. In contrast, existing studies have typically investigated similar phenomena within specific constructions, e.g., verb phrase ellipsis~\cite{Schafer2021-ga} and relativizer~\cite{Jaeger2007-wq} in English.
We specifically explored two key questions: (i) To what extent can native speakers agree on ellipsis judgments and their rationale in a natural, balanced sample of text?
(ii) How accurately can human ellipsis judgments be predicted by recent NLP approaches? 

\begin{figure}[t]
\centering
\includegraphics[width=7.5cm]{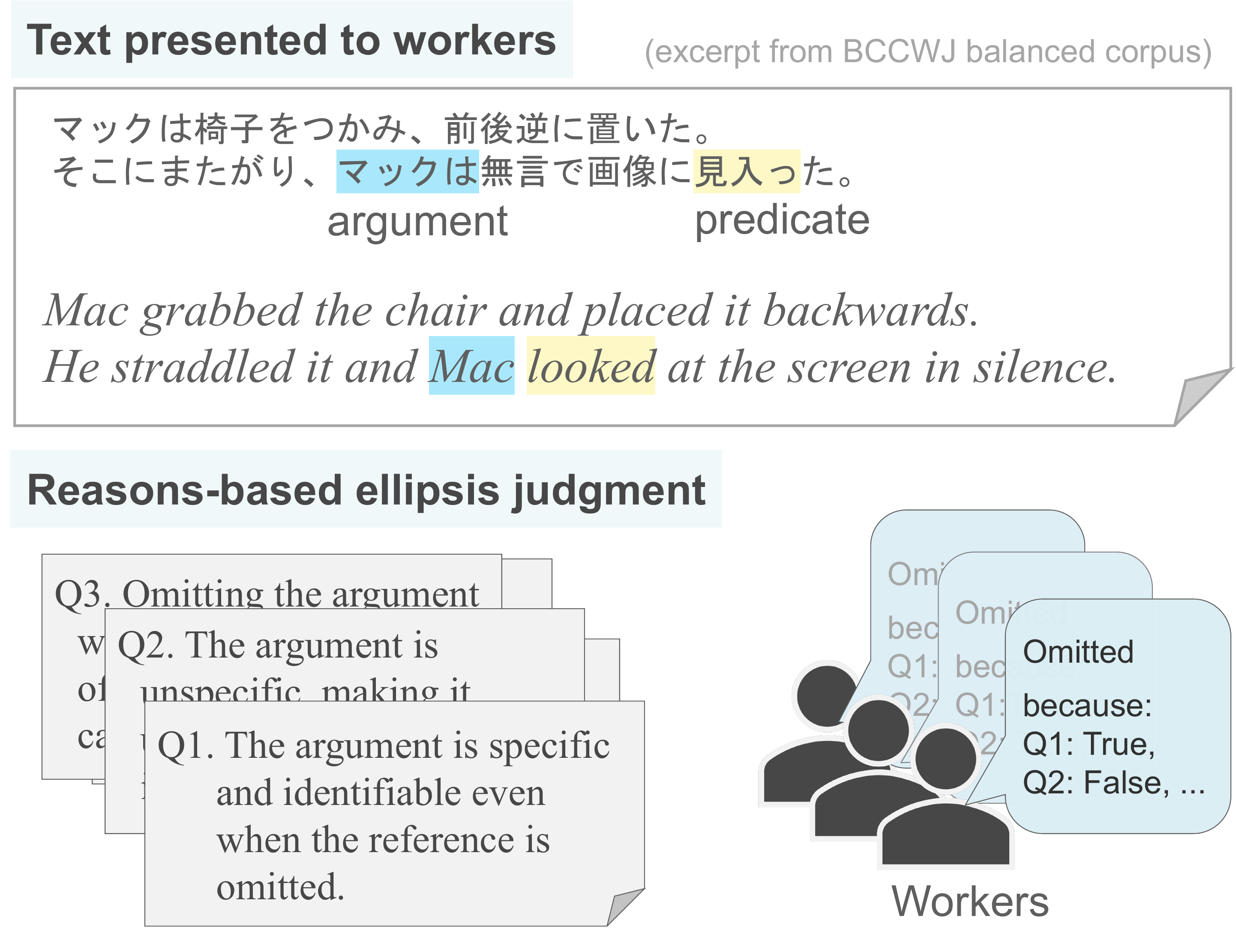}
\caption{Illustration of the argument ellipsis judgment annotation. In addition to asking for the final judgments, we also asked workers to answer the questions corresponding to the linguistic factors that potentially influence their judgments.}
\label{fig:overview}
\end{figure}

To achieve these objectives, we first created a dataset that determines whether and why a particular argument should be omitted.\footnote{Our data and codes are available at \url{https://github.com/cl-tohoku/JAOJ}.}
This dataset was built with over 2,000 data points extracted from a large, balanced Japanese corpus~\citep{maekawa2014balanced} and our newly designed annotation protocol and tool (Figure~\ref{fig:overview}).
Through creating our dataset, we found a general consensus in human judgments among the annotators, and the degree of this agreement varied depending on the associated linguistic factors. 
Furthermore, our annotations exhibited a clear parallel with the ellipsis judgments in the original corpus, suggesting a shared consensus about these judgments between the author (of the corpus text) and the readers (i.e., workers).

\begin{figure*}[t]
\centering
\includegraphics[width=\textwidth]{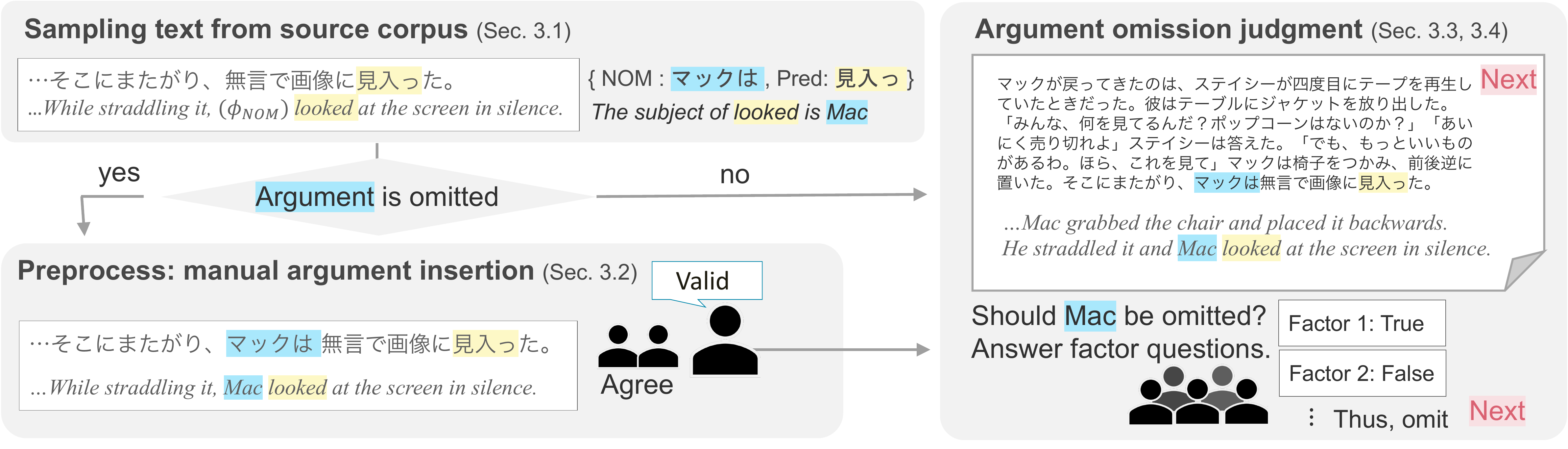}
\caption{Annotation procedure for our ellipsis judgment task.}
\label{fig:ann_process.pdf}
\end{figure*}

To examine question (ii), we benchmarked the performance of language models (LMs) on the ellipsis judgment task, including BERT, Japanese LMs, and GPT-4~\citep{devlin-etal-2019-bert, openai2023gpt4}.
Experimental results showed that the models struggled with certain types of ellipsis judgments, highlighting the human-LMs gap in discourse processing in the realm of ellipsis.


\section{Japanese Argument Ellipsis}\label{sec:Japanese_arg_ellipsis}
In Japanese, every argument of a predicate can potentially be omitted depending on the context:
\vspace{-0.5em}
\ex<tabeta>
\begingl
\gla  \colorbox{cyan!30}{Kare-wa} m\^o \colorbox{cyan!30}{kare-no ringo-wo} \colorbox{yellow!30}{tabe}-ta. //
\glb \textcolor{gray}{He-\texttt{TOP}} \textcolor{gray}{already} \textcolor{gray}{his\_apple-\texttt{ACC}} \textcolor{darkgray}{ate.} //
\glft `\colorbox{cyan!30}{He} already \colorbox{yellow!30}{ate} \colorbox{cyan!30}{his apple}.' //
\endgl
\xe
\vspace{-3em}
\ex<nai>
\begingl
\gla Kare-no ringo-wa nai. (\colorbox{cyan!30}{Kare-wa}) m\^o (\colorbox{cyan!30}{kare-no ringo-wo}) \colorbox{yellow!30}{tabe}-ta-node. //
\glb \textcolor{gray}{His} \textcolor{gray}{apple-\texttt{TOP}} \textcolor{gray}{there\_is\_not.} \textcolor{gray}{(He--\texttt{TOP})} \textcolor{gray}{already} \textcolor{gray}{(his\_apple-\texttt{ACC})} \textcolor{gray}{eat-\texttt{PAST}-\texttt{REASON}.} //
\glft `He doesn't have his apple. The reason is that (\colorbox{cyan!30}{he}) already \colorbox{yellow!30}{ate} (\colorbox{cyan!30}{his apple}).' //
\vspace{-0.5em}
\endgl
\xe

\noindent
For example, in Example (\getref{tabeta}), the subject (\textit{he}) and object (\textit{his apple}) of the predicate (\textit{eat}) are hardly omitted without context, but these would be omitted with a particular context as in Example~(\getref{nai}) considering the redundancy.
\noindent
The question is the decision criteria behind such ellipsis judgments.
Note that an argument ellipsis occurs in a chunk-level structure. 
Thus, when the phrase \textit{ringo-wo} (\textit{apple}) is dropped in Example~(\getref{nai}), its modifier \textit{kare-no} (\textit{his}) should also be omitted; this rule is applied to our dataset throughout this study.

Notably, our scope is beyond the typical focus of computational linguistics, for example, a syntactic reduction of relativizer (i.e., ``\textit{that}'')~\cite{Jaeger2007-wq}, in the sense that we commit more to the strategy of deciding what semantic information, i.e., who did what for whom, to say (or not to say) beyond the omission of supplemental, functional words/markers~\cite{Lee2006-kq}.
This broader scope makes it necessary for us to first break down the problem according to the associated linguistic factors (\S\ref{subsec:factors}), and hopefully, our systematic analysis will pose an elaborated question of which type of ellipsis can be explained by a particular approach, e.g., information-theoretic measurement~\cite{Jaeger2007-wq,meister-etal-2021-revisiting}.

\section{Annotation Task Design}
\label{sec:annotation_design}

\paragraph{Annotation Overview:}
Figure~\ref{fig:ann_process.pdf} shows data preprocessing (left part) and the annotation task (right part).
We first collected sentences potentially involving argument ellipsis from the balanced corpus (\S\ref{subsec:data} and \S\ref{subsec:preprocess}). 
Then, workers were provided with a sentence along with the information of which \colorbox{yellow!30}{predicate (\textit{looked})} and \colorbox{cyan!30}{argument (\textit{Mac})} are targeted and the preceding document-level context. 
Here, the workers were asked to infer whether the specified argument should be omitted or not along with its potential reason (\S\ref{subsec:factors} and \S\ref{subsec:factor_mapping}). 
The text was annotated using our newly designed tool for this task (\S\ref{subsec:tool}).


\subsection{Source Texts}\label{subsec:data}
\paragraph{Corpus:}
\textsc{BCCWJ-DepParaPas}~\citep{maekawa2014balanced}, a widely-used Japanese balanced corpus with a predicate-argument structure (PAS) annotation, was used to create instances for asking ellipsis preferences. 
We executed annotation using 32 documents from the book domain of this corpus.
The PAS annotation enabled us to identify the originally omitted arguments in the corpus; these can be used for collecting data points potentially favoring omission.
We targeted nominative (\texttt{NOM}), accusative (\texttt{ACC}), and dative (\texttt{DAT}) arguments that were annotated in this corpus.
Among the 17,481 arguments, including those omitted in the corpus, we finally used 2,373 arguments given the one-argument-per-sentence restriction (\S\ref{subsec:tool}).
The average length of the document-level contexts for each argument is 2,990 characters, and the average length of arguments is 6.25 characters.
Further details of the data sampling procedure are provided in Appendix~\ref{sec:appendix_sampling}.

\begin{table}[t]
\centering
\tabcolsep 3pt
\small
\begin{tabular}{lrrrrrr}
\toprule
& \multicolumn{3}{c}{Original} & \multicolumn{3}{c}{Sampled} \\
\cmidrule(lr){2-4} \cmidrule(lr){5-7}
Case & \#Args & Omit &  Insert & \#Args & Omit & Insert\\
\midrule
NOM & 10,537 & 46.2\%	& 53.8\% & 1,868 & 50.6\% &  49.4\% \\
ACC & 4,576 & 22.7\% & 77.3\% & 348 & 18.4\%  & 81.6\%  \\
DAT & 2,368 & 27.6\% & 72.4\% & 157 & 24.2\% & 75.8\%  \\
\midrule
Total & 17,481 & 37.5\% & 62.5\% & 2,373 & 44.1\% & 55.9\%  \\
\bottomrule
\end{tabular}
\caption{Omission rate of each grammatical case in the original and sampled corpus.}%
\label{table:cm_distribution}
\end{table}

\paragraph{Omission Statistics:}
Table~\ref{table:cm_distribution} shows the frequency of argument omission in the corpus.
The ``Original'' denotes the statistics of \textsc{BCCWJ-DepParaPas}, and the ``Sampled'' denotes the arguments targeted in this study.
Omission is not a rare phenomenon in a natural corpus; specifically, \texttt{NOM} argument is omitted as frequently as 
around 50\%.
The sampled arguments also reflect similar statistics to those in the original corpus.

\paragraph{Data Presented to Workers:}
The right part of Figure~\ref{fig:ann_process.pdf} shows the data provided to the workers.
The arguments were either present or omitted in the original sentence, and we asked the workers to make an ellipsis judgment---\textit{to drop or not to drop}---by showing a sentence with the target argument always \textbf{inserted}.
In other words, the workers were not provided with information regarding whether or not the argument was initially omitted in the corpus.
This protocol was employed to avoid any potential bias in their decision-making process.
An additional preprocessing was required to naturally insert the originally omitted arguments into the sentence (\S\ref{subsec:preprocess}).
Further details on the data annotation interface are explained in \S\ref{subsec:tool}.

\subsection{Preprocess: Argument Insertion}
\label{subsec:preprocess}
The left part of Figure~\ref{fig:ann_process.pdf} shows an overview of the argument insertion procedure, which is needed in advance to implement our annotation scheme.
To insert the originally omitted arguments into the sentence, a pair of workers, who were linguistics graduate students and not involved in the next ellipsis judgment task (\S\ref{subsec:factors}), were asked to determine both the appropriate surface form and the position of the argument to be filled in the sentence. Such a task of positioning is necessary due to the flexible nature of word-order in Japanese sentences.
If the two workers provided different answers, they were required to decide on the most plausible one. 
Subsequently, an additional worker further checked the data to exclude apparently collapsed sentences due to this insertion step.



\lingset{aboveglftskip=-.2ex,interpartskip=\baselineskip}

\begin{table*}[t]
\centering
\tabcolsep 4pt
\scriptsize
\renewcommand{\arraystretch}{1}
\begin{tabular}{p{3.4cm}p{10cm}p{1.0cm}} 
\toprule
Factor and Question & Example & Answer 
\\
\cmidrule(lr){1-1} \cmidrule(lr){2-2} \cmidrule(lr){3-3} 
\multirow{1}{3.6cm}{1. \textbf{Identifiability}: \\
The argument is specific and identifiable even when the reference is omitted.
}
& 
  \begingl
  \gla  Ringo-to mikan-ga aru. Saru-wa (\colorbox{cyan!30}{ry\^oh\^o-wo}) \underline{\colorbox{yellow!30}{taberu}}-koto-ga dekiru.//
  \glb Apple-\texttt{AND} orange-\texttt{NOM} be.  Monky-\texttt{TOP} (both-\texttt{ACC}) eating-\texttt{NOM} can\_do. //
  \glft `There are apples and oranges. The monkey can \colorbox{yellow!30}{\underline{eat}} (\colorbox{cyan!30}{both}).' //
  \endgl
& False 
\\
\cmidrule(lr){1-1} \cmidrule(lr){2-2} \cmidrule(lr){3-3} 
\multirow{2}{3.6cm}{2. \textbf{Specificity}: \\
The unidentifiable argument should be specified (True) or unnecessary to specify in the given context (False).
}
& \begingl
  \gla  Seifu-wa (\colorbox{cyan!30}{seifu-ga}) taisaku-wo \underline{\colorbox{yellow!30}{kento-suru}}-tame-ni iinkai-wo settisi-ta。 //
  \glb government-\texttt{TOP} (government-\texttt{NOM}) measure-\texttt{ACC} consider-\texttt{FOR} committee-\texttt{ACC} set\_up-\texttt{PAST}. //
  \glft `The government has set up a committee as (\colorbox{cyan!30}{the government}) \colorbox{yellow!30}{\underline{consider}} measures.' (government vs. comittee for the subject of \underline{\textit{consider}}) //
  \endgl
& False 
\\
\cmidrule(lr){1-1} \cmidrule(lr){2-2} \cmidrule(lr){3-3} 
\multirow{5}{3.6cm}{3. \textbf{Connotation}:\\
\textit{a}. Omitting the argument would result in different connotations. \\
$\;$ \\
(Only when 3\textit{a} is true) \\
\textit{b}. Connotation incurred by omission should be plausible given the context.\footref{foot:connotation}
}
& & \\
& & True \\
& \begingl
  \gla  (\colorbox{cyan!30}{Watashi-ga}) choshoku-wo \underline{\colorbox{yellow!30}{tabere-ba}} yokat-ta。 //
  \glb  (I-\texttt{NOM}) breakfast-\texttt{ACC} eat-\texttt{IF} was\_good.//
  \glft `(\colorbox{cyan!30}{No one else but I}) should have \colorbox{yellow!30}{\underline{eaten}} the breakfast.' \textcolor{gray}{(*Only the author (I) appears in this context)} //
  \endgl
& \\ 
& & True* \\
& & \\
\cmidrule(lr){1-1} \cmidrule(lr){2-2} \cmidrule(lr){3-3} 
\multirow{2}{3.6cm}{4. \textbf{Grammaticality}: \\
\textit{a}. Including the argument would render the sentence ungrammatical or redundant.}
& \begingl
  \gla  Mac-wa isu-wo tsukami, (\colorbox{cyan!30}{Mac-wa}) gyaku-ni \underline{\colorbox{yellow!30}{oi}}-ta.//
  \glb Mac-\texttt{TOP} chair-\texttt{ACC} grab-\texttt{AND}, (Mac-\texttt{TOP}) backwards put-\texttt{PAST}. //
  \glft `Mac grabbed a chair and (\colorbox{cyan!30}{he}) \colorbox{yellow!30}{\underline{set}} it backwards.' //
  \endgl
& True 
\\
\cmidrule(lr){2-2} \cmidrule(lr){3-3} 
\textit{b}. Omitting the argument would render the sentence ungrammatical or unnatural.
& \begingl
  \gla  Jiken-no kizi-wo sakusei-suru-tame, higaisha-ni (\colorbox{cyan!30}{hanasi-wo}) \underline{\colorbox{yellow!30}{k\^i}}-ta。//
  \glb incident-\texttt{OF} article-\texttt{ACC} create-\texttt{FOR}, victim-\texttt{DAT} (story-\texttt{ACC}) heard. //
  \glft `I \underline{\colorbox{yellow!30}{heard}} ({\colorbox{cyan!30}{the story}}) from  the victim to create an article about the incident.' //
  \endgl
& True 
\\
\cmidrule(lr){1-1} \cmidrule(lr){2-2} \cmidrule(lr){3-3} 
\multirow{2}{3.6cm}{5. \textbf{Miscellaneous preferences}: \\
Omitting/Inserting the argument enhances the naturalness of the sentence, although not necessary.}
& \begingl
  \gla    (\colorbox{cyan!30}{Watashi-wa}) wakai-koro, America-ni \underline{\colorbox{yellow!30}{ry\^ugaku-shi}}-te-ita。 \label{ex:npp}//
  \glb (I-\texttt{TOP}) when\_young, the~U.S.-\texttt{DAT} studied\_abroad. //
  \glft `(I) \colorbox{yellow!30}{\underline{studied}} abroad in the U.S. when (\colorbox{cyan!30}{I was}) young.' //
  \endgl
& Could go either way 
\\
\bottomrule
\end{tabular} 
\caption{Five factors found in the preliminary observation. We created the inquiry corresponding to each of these factors to ask workers the reason for their judgments.
The \textit{Answer} on the right side indicates whether the example sentence is True or False in response to the question about that specific factor.}
\label{table:ann_qa}
\end{table*}

\subsection{Factor-level Questions}
\label{subsec:factors}
One of our objectives is to explore the underlying rationale influencing human judgments of argument ellipsis. 
To achieve this aim, instead of enquiring about the ellipsis decision directly, we asked workers a series of questions based on the potential linguistic factors related to ellipsis judgment. 
The final judgment of a worker was subsequently derived from the responses to these questions.
The preprocessing and ellipsis judgment parts were performed by different annotators.

\paragraph{Factors:}
To investigate the linguistic factors associated with ellipsis judgments, the authors of this study conducted a trial annotation with small hold-out data. 
Then, we listed the following five factors by exploring potential reasons for cases in which the authors' choices aligned based on Japanese grammar~\citep{tsujimura2013introduction}.
Table~\ref{table:ann_qa} shows examples for each factor.

\begin{description}
  \setlength{\parskip}{0cm}
  \setlength{\itemsep}{0.1cm}
    \item[1. Identifiability:] 
    If a referent of the argument cannot be uniquely identified due to omission, it can potentially necessitate its explicit mention. This can also be seen as the difficulty of zero-anaphora resolution~\citep{iida2007zero}.
    \item[2. Specificity:] 
    If an unidentifiable argument needs to be specified to comprehend the text, it should be realized. 
    Otherwise, if a context does not require uniquely specifying the argument entity, one can leave it unspecified by omitting it~\citep{dipper2007information}.
    \item[3. Connotation:] 
    Affixed particles to the argument can serve a diverse range of connotative functions, such as topicalization and emphasizing contrasts or exclusivity. The argument must be retained if its omission can lead to the irrevocable loss of subtle connotations~\citep{kenkyukai2008gendai}.\footnote{Practically, there might be a case wherein a worker cannot guess the preferable decision only from the context when Factor 3 is true (ellipsis affects the connotation). At least during our annotation, such a case rarely occurs; therefore, we just ignored these instances herein.\label{foot:connotation}}
    \item[4. Grammaticality:] 
    If the realization or omission of the argument considerably disrupts the syntactic fluency of the sentence, it is undesirable to do so.
    \item[5. Miscellaneous preferences:] 
    When the aforementioned factors are inadequate for judgment, a more favorable one is selected based on naturalness.
\end{description}

\paragraph{Constraints and Preferences:}
We presumptively associated Factors 1-4 with a type of linguistic constraint and Factor 5 with preferences, respectively.
In other words, different ellipsis judgments associated with Factors 1-4 can directly affect the information contents communicated between the writer and readers (Factors 1-3) or the accuracy of communication (Factor 4). In this sense, these factors were constrained to make the communication disambiguated and clear.
In contrast, Factor 5 can be a type of subjective preference related to the naturalness of the text (e.g., preference toward salience and the efficacy/redundancy of communication).
We tentatively introduced this categorization in the analyses and experiments (\S\ref{sec:experiment} and \S\ref{sec:ex2}).
Specifically, we referred to the ellipsis judgments associated with Factors 1-4 as ``\textit{hard-omission and insertion}'' (HO and HI) and those with Factor 5 as ``\textit{soft-omission and insertion}'' (SO and SI), respectively (see \S\ref{subsec:factor_mapping} for details). 
This distinction will generally be tied with the bi-dimensional pressures underlying communication: informativity (constraints) and complexity (preferences) (\citealt{I_Cancho2003-kd,Piantadosi2012-de,Frank2012-uu,Kirby2015-kz,Gibson2019-oe,Xu2020-at,Hahn2020-dq}; \textit{inter alia}).


\subsection{From Factor-level Answers to Ellipsis Judgments}
\label{subsec:factor_mapping}
\begin{figure}[t]
    \centering
    \includegraphics[width=7.5cm]{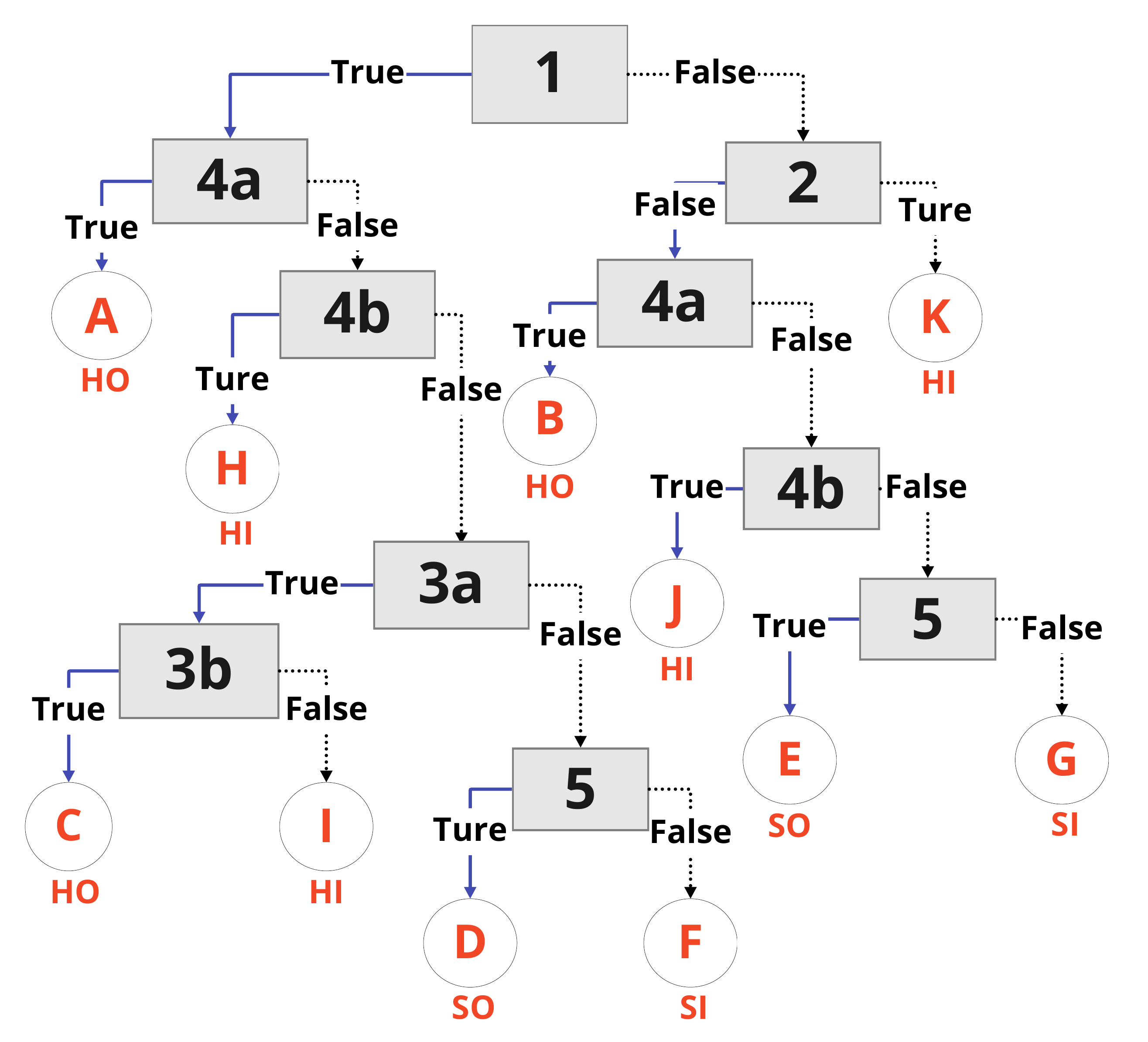}
    \caption{The decision tree used in the annotation process. Each number denotes the factor ID. The validity of the tree was confirmed during the preliminary tasks among the authors and the instructions provided to the workers.    
    }
    \label{fig:appendix_decision_tree}
\end{figure}

The workers were presented with the factor-level questions first and were asked to arrive at a decision on whether to omit the argument based on their answers.
Specifically, the answers to the factor-level questions were mapped to the final ellipsis decision based on the decision tree shown in Figure~\ref{fig:appendix_decision_tree}.
The node label in Figure~\ref{fig:appendix_decision_tree} corresponds to the factor ID in Table~\ref{table:ann_qa}.
For example, if a worker answers \textit{False} for \textit{Factor 1} (identifiability) and \textit{True} for \textit{Factor 2} (specificity), the argument should be inserted as it is necessary to comprehend the text and it cannot be complemented with the context information.


To reduce the load on the workers, the questions were asked in the top-to-bottom order shown in Figure~\ref{fig:appendix_decision_tree}, i.e., Factor~1 is the first question.
The worker finished answering when the answers could be mapped to the final ellipsis decision---whether it should be omitted or not.


\subsection{Annotation Interface}
\label{subsec:tool}
\paragraph{Interface:}
To make the annotation efficient and controlled, we developed a dedicated annotation tool.
Within this tool, the targeted \colorbox{cyan!30}{argument} and \colorbox{yellow!30}{predicate} are highlighted, and the worker answers the factor-level questions (\S\ref{subsec:factors}) on the targeted argument with the displayed context.
When workers press the ``Next'' button on the screen (right side of Figure~\ref{fig:ann_process.pdf}), the subsequent target predicate-argument pair and its context are appended after the current text. 
In other words, the targeted arguments are shifted in the order in which they appear from the beginning of the text, resulting in the workers reading the entire text incrementally as they proceed with the task.
This process is repeated until the end of one document.
Note that to simplify the task, a single argument-predicate pair is targeted in each sentence, as described in \S\ref{subsec:data}.
That is, when a sentence contains more than two arguments in a sentence, one to be worked on is randomly determined in advance.

\paragraph{Original Ellipsis Decision is not Leaked:}

As described in \S\ref{subsec:data} and \S\ref{subsec:preprocess},
the workers were not provided with information regarding whether the argument was initially omitted or not in the original corpus.
To ensure this but prevent the incorrect interpretation of context, after the workers answered the questions and pressed the ``Next'' button, the original sentence was disclosed as the actual prior context for the forthcoming target instances.
Therefore, we strictly prohibited the workers from revising their preceding judgments.

\section{Data Creation}
\label{sec:data}

\subsection{Workers}\label{workers}
The ellipsis judgment task was conducted by five native Japanese speakers who were university students.
Prior to the task, they participated in an instruction and training session where we ensured that they could achieve the expected judgments, which were predetermined by the authors of this study, in a hold-out small dataset.

\subsection{Data Statistics}
\label{subsec:statistics}


\paragraph{Label Distributions:}

\begin{table}[t]
\centering
\tabcolsep 3pt
\small
\begin{tabular}{crrrrr}
\toprule
&& \multicolumn{2}{c}{Omitted} &  \multicolumn{2}{c}{Inserted} \\
\cmidrule(lr){3-4}  \cmidrule(lr){5-6} 
Case & \#Args & Hard &  Soft &  Soft  &  Hard \\
\midrule

NOM & 1,868 & 35.2\% & 14.9\% & 11.5\% & 38.4\% \\
ACC & 348 & 13.0\% & 8.9\% & 18.0\% & 60.0\% \\
DAT & 157  & 13.0\% & 5.7\% & 8.6\% & 73.0\% \\
\midrule
Total & 2,373 & 30.4\% & 13.2\% & 11.5\% & 44.9\% \\

\bottomrule
\end{tabular}
\caption{Distribution of aggregated human ellipsis judgment labels for each grammatical case.}
\label{table:cm_ann_result}
\end{table}

To show the overall statistics, we aggregate the human judgment labels by the five workers for each data point as their median.
Here, we assume the original scale of HO$\succ$SO$\succ$SI$\succ$HI (see \S\ref{subsec:factors} for these categorizations).
Table~\ref{table:cm_ann_result} shows the label distribution in the collected data, where an overall trend that hard omissions/insertions more frequently occur than soft ones can be observed.
Moreover, the distributions varied depending on the grammatical case types; \texttt{NOM} arguments were omitted more often (50.1\%=35.2\%+14.9\%) than the other cases.
Note that such an omission rate was similar to that in the corpus (Table~\ref{table:cm_distribution}).

\paragraph{Inter-worker Agreement:}
We calculated Krippendorff's alpha, the agreement measure for ordinal labels, for our five workers, resulting in 0.87. 
This indicates that the workers' judgments were generally consistent.\footnote{A value greater than 0.677 is considered a high agreement~\citep{Krippendorff2004}
.}
To investigate the label-wise inter-worker agreement,
we calculate a one-vs-one $F_1$ score for every pair of two workers and consider an average of the scores.
The first line in Table~\ref{table:lm_result} shows the resulting scores.
As suggested in \S\ref{subsec:factors}, the constraint-based judgments (Factors 1--4) show comparatively higher scores of 74.6\% for HO and 87.2\% for HI.
The scores of preference-based judgments (Factor 5) were considerably low (35.9\% and 39.3\%), suggesting that these categories have some degree of freedom in the choice.

\begin{figure}[t]
\centering
\includegraphics[width=7.7cm]{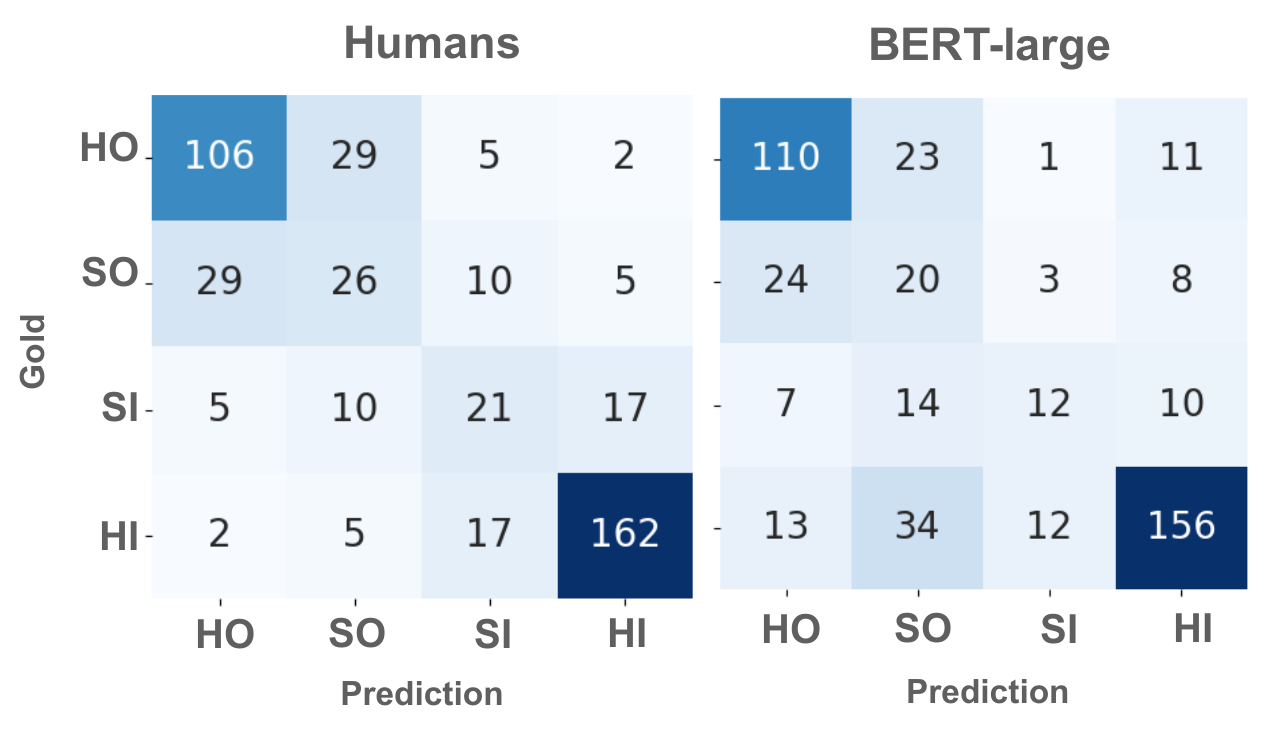}
\caption{Confusion matrix of humans (left side) and BERT-large (right side) in validation data. Note that the numbers of instances are based on the average of one-vs-one evaluations for humans and based on the comparison between the system's predictions and the aggregated labels of humans for BERT$_\mathrm{L}$.}
\label{fig:cm_human_lm}
\end{figure}

\paragraph{Inter-worker Confusion Matrix:}
The left part of Figure~\ref{fig:cm_human_lm} shows the confusion matrix on the one-vs-one evaluation. The number of instances was averaged by the number of trials. 
We observed that the SO/SI judgments had a larger variance than the HO/HI judgments.
Moreover, the different label choices among the workers basically occurred between adjacent labels (e.g., HO and SO, or SO and SI) and rarely in between distant labels (e.g., HO and HI, or SO and HI).

\paragraph{Agreement with the Original Corpus:}
We also compared the workers' judgments with those made in the original corpus by converting four categories of  HO, HI, SO, and HI into the binary decisions between omission and insertion.
The agreement score was calculated in terms of the accuracy of these binary judgments in the original text and the aggregated label, resulting in 97.0\%. This score suggests that the judgments for argument omission were consistent between a writer (authors of the corpus texts) and readers (annotators).

\paragraph{Factor-level Statistics:}
Table~\ref{table:factor_distribution} shows the distribution of linguistic factors associated with the worker judgments.
We confirmed that no specific linguistic factor dominated the worker judgments, but the associated factors spanned the listed multiple factors.
Note that the associated factor for each argument was aggregated on a majority voting basis (see \S\ref{sec:ex2}).

\begin{table}[t]
\centering
\small
\begin{tabular}{lrrrrr}
\toprule
& \multicolumn{5}{c}{\#Instances} \\
\cmidrule(lr){2-6} 
Label &  \multicolumn{1}{c}{1} & \multicolumn{1}{c}{2} & \multicolumn{1}{c}{4\textit{a}} & \multicolumn{1}{c}{4\textit{b}} & \multicolumn{1}{c}{5} \\
\cmidrule(lr){1-1} \cmidrule(lr){2-2} \cmidrule(lr){3-3}  \cmidrule(lr){4-4} \cmidrule(lr){5-5} \cmidrule(lr){6-6}%
True  & 1,419& 736 &  119 &  25 & 295 \\
False &  940 & 704 & 1,322 & 914 & 273 \\
\bottomrule
\end{tabular} 
\caption{Statistics of factor-level questions.}
\label{table:factor_distribution}
\end{table}

\section{Experiment: Predicting Human Ellipsis Judgments}
\label{sec:experiment}

We benchmark ellipsis judgment prediction models powered by standard natural language processing (NLP) approaches: (i) zero-shot prompting with large LMs, such as GPT-4~\citep{openai2023gpt4} and the Japanese LLM, Swallow (\S\ref{subsec:models}), and (ii) finetuning with BERT~\citep{devlin-etal-2019-bert}.
Note that pursuing a high-performance model for this task is not the objective of this paper; rather, we use these straightforward baselines to introduce our task.


\subsection{Task}
\label{sec:ex1_settings}
\paragraph{Task Formulation:}

We formulate the task involving the prediction of the aggregated judgment label of five workers (computed as in \S\ref{subsec:statistics}), i.e., one of the following four categories: HO, SO, SI, and HI. 
We employ the categorical classification as our initial foray.
In a future study, we will introduce an ordinal scale to the models and metrics will be our future work.
We used $F_1$ as the evaluation metric herein.

\paragraph{Data Split:}
We partitioned the data points into the training (1,459 arguments), validation (456), and test (458) sets.
This partitioning was conducted at the document level, thus ensuring that data points across different folds do not share the context from the same document.
We confirmed that the label distribution for each split was comparable.
\subsection{Models}
\label{subsec:models}
\begin{figure}
    \centering
    \includegraphics[width=7.5cm]{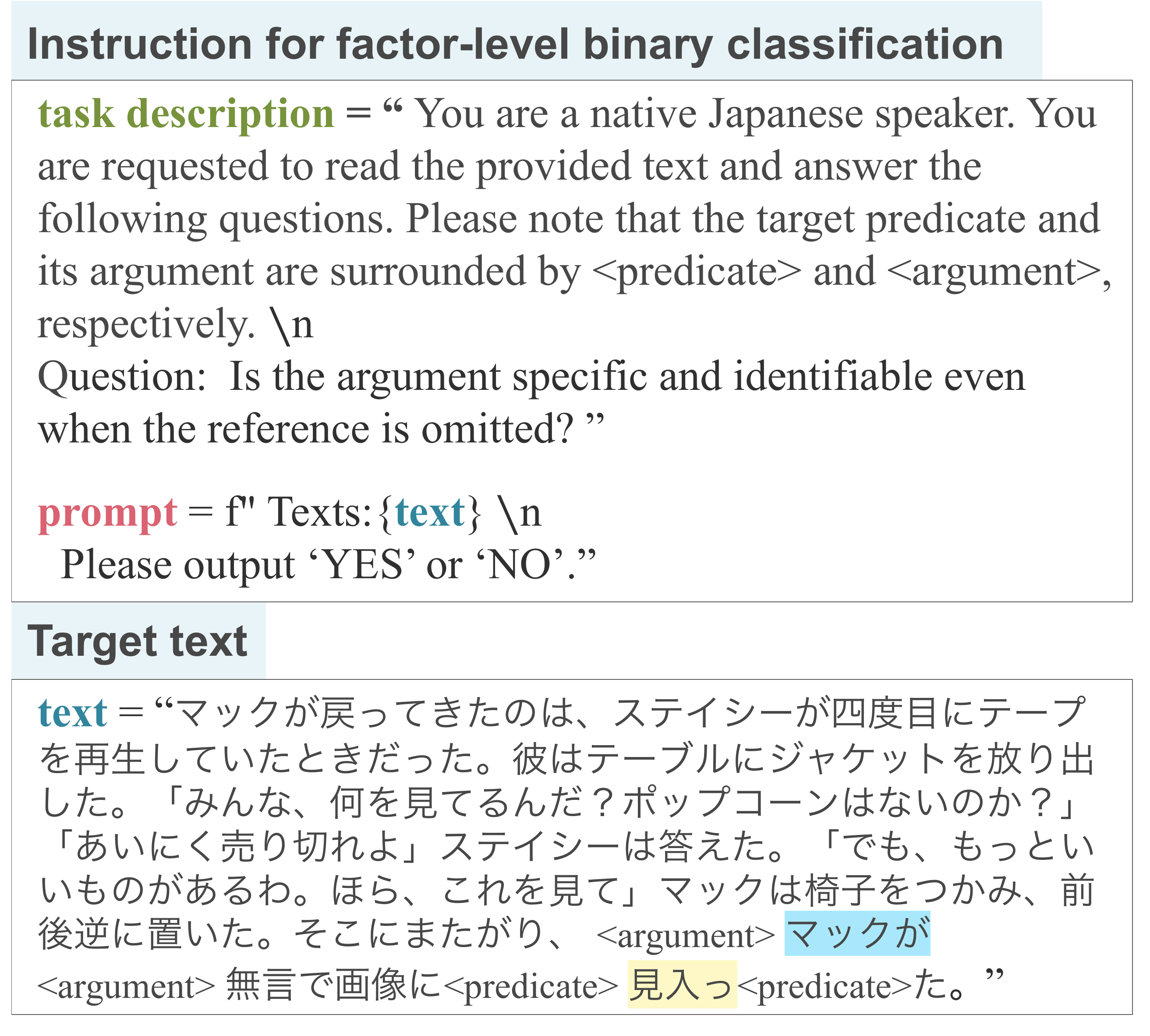}
    \caption{Prompt example for the GPT models. The targeted predicate and argument are surrounded by special tags (\texttt{<argument>} or \texttt{<predicate>}). An explanation of the targeted factor and an example illustrated in the annotation manual are also provided to the model.}
    \label{fig:prompt}
\end{figure}

\paragraph{GPT-3.5 and GPT-4:}
GPT-3.5 and GPT-4 are multilingual models and have sufficient ability to use Japanese \citep{Kasai2023-tm}. 
We obtain their argument ellipsis judgments by posing factor-level questions to them, utilizing a virtually identical protocol to that designed for human workers (\S\ref{sec:annotation_design}), which enables a direct comparison between humans and models.
Figure~\ref{fig:prompt} shows the prompt template for a particular factor-level question, where the description of ``Question'' is changed depending on the corresponding factor-level question.
The target context and predicate-argument pair are fed into the models, following the question.\footnote{The context is limited to 512 subwords, considering the BERT's maximum input length. We randomly chose the answer label when the models did not yield a specific answer.}
The final ellipsis judgment is induced from the responses to all factor-level questions in the same manner as explained in \S\ref{sec:annotation_design}.

\paragraph{Japanese large LMs:}
We also test Japanese large LMs, specifically, instruction-tuned versions of Swallow LMs with 13B (Swallow$_\mathrm{S}$)\footnote{ \url{https://huggingface.co/tokyotech-llm/Swallow-13b-instruct-hf}} and 70B (Swallow$_\mathrm{L}$)\footnote{ \url{https://huggingface.co/tokyotech-llm/Swallow-70b-instruct-hf}} parameters, which are created by further pre-training of Llama 2 \citep{touvron2023llama} with Japanese data. 
We use the same prompting setting as GPT-3.5 and GPT-4 experiments.

\paragraph{BERT:}
As another simple approach, we fine-tune BERTs, where the input is the ``target text'' shown in Figure~\ref{fig:prompt} alone, and the training objective is simply handled as the four-class classification task.
We use BERT-base-japanese (BERT$_\mathrm{S}$)\footnote{\url{https://huggingface.co/cl-tohoku/bert-base-japanese-v2}} and BERT-large-japanese (BERT$_\mathrm{L}$).\footnote{\url{https://huggingface.co/cl-tohoku/bert-large-japanese}}
We fine-tune models with five different seeds and report the standard deviation as well as the average of their performance.
More detailed settings of input sequences and hyperparameters are described in Appendix~\ref{sec:appendix_hypara}.

\paragraph{Baselines:}
We also examine two simple baselines: (i) \textit{Rand} that predicts labels randomly, and (ii) \textit{Rand+} that assigns the HI label for the \texttt{DAT} and \texttt{ACC} arguments, which are the majority label for these grammatical cases, and randomly assigns the decision label for the \texttt{NOM} argument.

\begin{table}[t]
\centering
\small
\tabcolsep  1pt
\begin{tabular}{lccccc}
\toprule
& Avg. & \multicolumn{2}{c}{Omitted} &  \multicolumn{2}{c}{Inserted} \\
 \cmidrule(lr){3-4}  \cmidrule(lr){5-6}  
Model & Macro  & Hard & Soft & Soft & Hard \\
\midrule
Humans & 59.3\tiny{$\pm{2.8}$} & 74.6\tiny{$\pm{2.5}$} & 35.9\tiny{$\pm{9.7}$} & 39.3\tiny{$\pm{6.0}$} & 87.2 \tiny{$\pm{2.2}$}\\
\midrule
$\text{GPT-3.5}$ & 21.4 & 17.5 & 	4.2  & 15.1 & 48.9  \\
$\text{GPT-4}$ & 25.2 & 	32.7 & 	11.6 & 	9.2 & 47.2  \\
\midrule
$\text{Swallow}\tiny{_\text{S}}$ & 21.7 & 40.8 & 7.0 & 	3.8 & 35.2   \\
$\text{Swallow}\tiny{_\text{L}}$ & 22.7 & 	36.7 & 	10.5 & 	7.6 & 35.9   \\
\midrule
$\text{BERT}\tiny{_{\text{S}}}$ & 51.0\tiny{$\pm{0.6}$} & 69.5\tiny{$\pm{0.5}$} & 24.0\tiny{$\pm{2.0}$} & 32.3\tiny{$\pm{2.3}$} & 78.1\tiny{$\pm{0.6}$} \\
$\text{BERT}\tiny{_{\text{L}}}$ & 50.2\tiny{$\pm{4.3}$} & 76.0\tiny{$\pm{4.7}$} & 29.6\tiny{$\pm{3.7}$} & 10.9\tiny{$\pm{5.0}$} & 84.2\tiny{$\pm{4.2}$}\\
\midrule
Rand  & 20.2 & 28.2 &	 11.4 & 10.8 & 30.3\\
Rand+ & 27.2 & 27.0 &	 22.5 &	11.8 & 47.7\\
\bottomrule
\end{tabular} 
\caption{Performance of human workers and the tested models.
The values of human and BERT models are the average and standard deviation of five workers/models' results.}
\label{table:lm_result}
\end{table}

\subsection{Results}
\label{sec:ex1_result}

Table~\ref{table:lm_result} shows the results of the four-class classification. 
As explained in \S\ref{subsec:statistics}, the human performance is the average $F_1$ score calculated through one-vs-one comparisons of each worker pair's predictions.
We observed that the GPT and Swallow models only achieved baseline-level performance, 
indicating that they cannot replicate human-like Japanese ellipsis judgments, at least within this prompting setting.

The fine-tuned BERTs exhibited better performance than the prompt-based models, although this difference itself was not surprising since BERTs were trained with human annotations, whereas GPTs and Swallows were under the zero-shot prompting setting.
BERT's performance was still inferior to humans in some aspects; predicting the soft preferences (SO/IO) was particularly challenging for the fine-tuned BERT compared to the human upper bound.
The right side of Figure~\ref{fig:cm_human_lm} shows the confusion matrix of the BERT$_\mathrm{L}$ model, which incorrectly predicted the distant labels than humans (e.g., HI v.s. SO).
In general, the scores for HO/HI consistently surpass those for SO/SI. This tendency was also observed in human judgments, although the performance gap is more significant for LM-based models.
In summary, all the tested models could not fully simulate human-like ellipsis judgments, which is a challenge for LMs.

\section{Model Analysis}
\label{sec:ex2}
In this section, we analyze the ability of LMs for ellipsis judgments from the factor-level perspective.

\paragraph{Task:}
To evaluate the difficulty faced by LMs in answering the factor-level questions, we report the binary classification performance for each factor listed in \S\ref{subsec:factors}.
Here, a model is provided with the same instance as the ellipsis judgment task, but the task is to predict whether the answer is \textit{True} or \textit{False} for each factor-level question.

\paragraph{Models:}
As in \S\ref{sec:experiment}, we examine zero-shot prompting with the GPT and Swallow models and finetuning with BERTs, but again the task is the factor-level binary classifications.
Note that the answers of GPTs and Swallows to the factor-level questions are exactly those used in \S\ref{sec:experiment} to induce the final ellipsis judgment.
Models are evaluated in terms of the macro $F_1$ of \textit{True}/\textit{False} labels.
For human performance, the same one-vs-one evaluation method described in \S\ref{subsec:statistics} is used.

\paragraph{Data:}\label{ex2:data}
We partitioned each data point into binary classification tasks of \textit{True}/\textit{False}, each of which corresponds to the question of a specific factor.
Here, we only used the data points where the gold labels could be determined via majority votes among the workers; this resulted in the creation of a different train/valid/eval split in a 3:1:1 ratio with the main experiment (\S\ref{sec:experiment}).
A very small number of samples were associated with Factor 3 (Table~\ref{table:factor_distribution}), thus this factor was excluded from the analysis. 
These procedures are detailed in Appendix~\ref{sec:appendix_finetuning}.


\paragraph{Results:}

\begin{table}[t]
 \centering
 \small
 \tabcolsep  1pt
 \begin{tabular}{lccccc}
 \toprule
 Model &  Fac.1 &  Fac.2 & Fac.4\textit{a} & Fac.4\textit{b} & Fac.5 \\
 \cmidrule(r){1-1} \cmidrule(lr){2-2} \cmidrule(lr){3-3}  \cmidrule(lr){4-4} \cmidrule(lr){5-5} \cmidrule(lr){6-6}%
 Human & 86.3\tiny{$\pm{1.3}$} & 64.2\tiny{$\pm{2.1}$} &80.5\tiny{$\pm{0.7}$}	& 71.1\tiny{$\pm{2.4}$} & 70.6\tiny{$\pm{2.4}$}\\
 \midrule
  $\text{GPT-3.5}$ & 44.0 & 39.4 & 49.8 & 43.9 & 35.3 \\
 $\text{GPT-4}$ & 53.2 & 29.5 & 51.2 & 33.8 & 48.0 \\
 \midrule
 $\text{Swallow}\tiny{_{\text{S}}}$ & 49.5 & 26.1 & 48.9 & 31.8 & 46.4 \\
 $\text{Swallow}\tiny{_{\text{L}}}$ & 48.8 & 31.3 & 49.1 & 36.3 & 51.4 \\
  \midrule
 $\text{BERT}_\text{S}$ & 85.5\tiny{$\pm{0.4}$} & 53.3\tiny{$\pm{1.6}$} & 76.8\tiny{$\pm{1.6}$} & 61.6\tiny{$\pm{1.8}$} & 58.8\tiny{$\pm{2.1}$} 	\\
 $\text{BERT}_\text{L}$ & 84.6\tiny{$\pm{0.5}$}  & 52.5\tiny{$\pm{1.8}$} & 70.0\tiny{$\pm{4.4}$} & 56.0\tiny{$\pm{3.0}$}  & 58.8\tiny{$\pm{2.7}$}	\\
 \midrule
 Rand & 47.9 & 49.0 & 37.8 & 37.7 & 48.4 \\
 \bottomrule
 \end{tabular} 
 \caption{Macro $F_1$ scores of the factor-level binary classifications. The values of human and BERT models are the average and standard deviation of five workers/models' results. The values of Rand are influenced by the imbalance in the distribution of the labels in Table \ref{table:factor_distribution}.}
 \label{table:result_ex2}
\end{table}

Results are shown in Table \ref{table:result_ex2}.
Some parallels were observed in the performance differences across the factors between humans and BERTs. Factors 1 and 4a were relatively easy to solve, and Factors 2, 4b, and 5 were difficult for all humans and BERTs.
Specifically, BERTs exhibited nearly human-upper bound performance in Factor 1 (identifiability), whereas a relatively large BERT--human performance gap was observed in Factor 5 (preference).
These findings clarify that particular linguistic factors are associated with ellipsis judgments that are difficult for the models to predict.
Notably, the GPT and Swallow models still yielded nearly random baseline performance in almost all factors except Factor 4a.

\section{Related Work}\label{sec:related_work}
\subsection{Ellipsis in Computational Linguistics}
The scope of ellipsis studies spans various fields, including theoretical syntax, semantics, discourse, and psycholinguistics~\citep{merchant2001syntax,Van_Craenenbroeck2019-es}.
Our investigation will largely be aligned with psycholinguistic interests in ellipsis, aiming to unveil the biases associated with the ellipsis judgment or coreference resolution~\cite{Carminati2005-qm,michaelov-bergen-2022-language} during language processing; other perspectives are also involved in our listed linguistic factors though (\S\ref{subsec:factors}).
Some studies have tested specific hypotheses on ellipsis judgments, such as uniform information density~\citep{Jaeger2007-wq,Schafer2021-ga} hypothesis and centering theory~\citep{Walker1994-ah,Grosz1995-qe}, using controlled examples.
This study, in contrast, employs a complementary empirical approach of analyzing the types of ellipsis that occur frequently in the balanced corpus and predicting them from both linguistic and engineering perspectives.
We hope that our created resource facilitates further studies toward ellipsis judgment.

\subsection{Ellipsis in NLP}
In NLP, the zero anaphora resolution that automatically recovers the omitted argument/antecedent from its context has been actively explored~\citep{sasano-kurohashi-2011-discriminative,noauthor_undated-jn,Konno2021-lu}.
Complementary to such studies, our study focuses on a different problem---when the argument should be omitted and why based on its context.
This perspective is also important when writing/generating natural texts, achieving writing assistance, and computationally modeling human discourse processing.
Notably, the Japanese language we focused on is a popular choice in such ellipsis studies~\cite{Iida2007-fx, Shibata2018-en, Konno2021-lu}.

\subsection{Analyzing Discourse Processing of NLP Models}
Discourse is a challenging linguistic aspect to process computationally.
NLP models have been tested/probed from various discourse perspectives, and their hindered or non-human-like abilities have been typically reported~\citep{Bender2020-ds,Sakaguchi2021-bi,Upadhye2020-ed,Koto2021-rl,Schuster2022-uc,Fujihara2022-zz}.
The experiments conducted herein were also aimed in this direction and exhibited that predicting the soft preferences was particularly challenging for the models.

Additionally, the linguistic competence of large LMs, e.g., the ability to judge the acceptability of a sentence \citep{Dentella2023-yv}, has been tested via meta-linguistic prompting~\citep{Hu2023-gx}.
We also employed such a prompting approach in evaluating large LMs to align the evaluation scheme with humans and facilitate their direct performance comparison.
Nevertheless, \citet{Hu2023-gx} pointed out that the prompt-based method tends to underestimate the model's linguistic knowledge, and the probability-based evaluation is indeed another standard approach for quantifying the (context-dependent) sentence acceptability judgment ability of LMs~\cite{warstadt-etal-2020-blimp-benchmark,hu-etal-2020-systematic,sinha-etal-2023-language}.
Thus, benchmarking the model's ellipsis judgment preference through generative, probabilistic perspectives should be future work, although simple probability comparison might pose another potential limitation of simply avoiding a sentence containing new information (not identifiable in Factor 1, though) with low probability.

\section{Conclusion}
In this study, we examined the degree of consensus among native speakers on ellipsis judgments and their rationale within a naturally balanced text sample, motivated by the interest in human discourse processing and potential application to writing assistance. 
To this end, we have collected human annotations of whether and why a particular argument should be omitted across over 2,000 data points in a large balanced corpus. 
The collected data varied based on their associated linguistic factors.
We also examined the performance of the LM-based argument ellipsis judgment model and quantified the gap between the system's and human's judgments.
We hope that the annotation data we created serves as a foundational resource for future research toward argument omission.


\section{Limitations}
The annotation task was somewhat artificially designed compared with the human's real reading/writing activities.
Another approach, such as directly measuring human reading behavior (e.g., reading time), will be complementary to our study.
In addition, the number of workers involved in our annotation task was limited to five; thus, there might be some annotator biases in human gold judgment, although they were carefully instructed in trial sessions and showed high agreement, and we attempted to avoid these biases in workers' majors, academic years, and genders.
We also suspect that some data points, such as those with "soft" decisions, may particularly reflect workers' biases, such as their reading ability. 
Analyzing such biases themselves may provide additional insights into human language processing, and thus, this will be an interesting future research.

The adopted linguistic factors 1--5 and the course categorization of SO/HO/HI/SI were mainly based on the trial examinations with the authors of this paper and previous linguistic studies.
Quantitative support for this decision should have been preferred.
Furthermore, Factor 5 (miscellaneous preferences) was a somewhat vague category and required further categorization.
Nevertheless, this study isolated the data points with respect to potential linguistic factors.
This will enable researchers to conduct studies by focusing more on specific aspects (e.g., preference) of ellipsis.
For example, contrasting information-theoretic statistics with Japanese ellipsis preference will provide support for a particular hypothesis from a non-English perspective.

As for the experimental designs, the input for the models was limited to a particular token length (512 tokens).
This setting might have underestimated the model performance, although in most cases, the context was less than 512 tokens in length.
We also observed that GPT-3.5, GPT-4, and Swallow models yielded chance-level performances, which may be further improved by refining prompt designs or examining probability measurements.
Regardless, the experiments in this paper are positioned as a demonstration aimed at predicting human ellipsis judgments. Consequently, the improvement of the performance will be the subject of a future investigation.

\section{Ethics Statement}
We ensured the collected data did not contain any information that could identify an individual worker. The hourly wage of each worker was the standard salary of university students in their country.

\section{Acknowledgement}
This work was supported by JST CREST Grant Number JPMJCR20D2.

\section{Bibliographical References}
\bibliographystyle{lrec-coling2024-natbib}
\bibliography{custom}


\clearpage
\section*{Appendix}
\appendix

\section{Sampling Arguments from Corpus}\label{sec:appendix_sampling}
In this study, we focused on arguments corresponding to \texttt{NOM}, \texttt{DAT}, and \texttt{ACC} cases among the arguments appearing in the source corpus.
We also excluded the predicates that primarily serve a functional or grammatical role rather than conveying specific semantic content such as ``aru'' (be), ``naru'' (become), and ``yaru''
 (do) in advance. 
 
The original corpus sometimes contained multiple predicates within a single sentence.
In cases wherein a sentence had multiple predicates, the predicate-argument pair to be annotated was determined randomly to ensure that each annotated sentence had only one target argument.

\begin{table}[t]
\centering
\tabcolsep 4pt
\begin{tabular}{rrrrrr}
\toprule
 & & \multicolumn{2}{c}{Omit} &  \multicolumn{2}{c}{Insert} \\
\cmidrule(lr){3-4}  \cmidrule(lr){5-6} 
 Split & \#Args. & Hard & Soft  & Soft & Hard \\
\midrule
Train &1,459 & 29.5\% & 15.2\% & 11.4\% & 43.9\%  \\
Valid &456 & 31.6\%  & 8.1\%  & 14.3\%  & 46.1\%\\ 
Test &458 & 31.6\%  & 11.8\%  & 9.6\% & 47.1\% \\  
\midrule
All & 2,373 & 30.3\% & 13.2\% & 11.6\% & 44.9\% \\
\bottomrule
\end{tabular}
\caption{Label distribution of the dataset.}
\label{table:class_distribution}
\end{table}

\section{Label Distribution of Dataset}\label{sec:appendix_data}
The label distribution of the dataset used in the experiment in \S\ref{sec:experiment} is shown in Table~\ref{table:class_distribution}.

\section{Hyperparameter and Model Settings}\label{sec:appendix_hypara}
\begin{table}
\centering
\small
\tabcolsep  2pt
\begin{tabular}{ll|ccccc}
\toprule
&& \multicolumn{5}{c}{Factors} \\
Archtecture & Params & 1 & 2 & 4\textit{a} & 4\textit{b} & 5 \\
\midrule
$\text{BERT}_\text{S}$ & batch size & 8 & 8 & 16 & 8 & 8 \\
 & learning rate & 5e-5 & 5e-7 & 3e-5 & 2e-5 & 3e-5\\
\midrule
$\text{BERT}_\text{L}$ & batch size & 16 & 8 & 16 & 16 & 16 \\
 & learning rate & 2e-5 & 5e-6 & 2e-5 & 5e-5 & 2e-5 \\

\bottomrule
\end{tabular} 
\caption{Batch sizes and learning rates for \S\ref{sec:ex2}. }\label{tbl:appndix_hypara}
\end{table}

All BERT-based classification models were trained on NVIDIA RTX A6000.
Early stopping was applied based on the loss of training data.
All other parameters not specified in this section followed the default values of the \texttt{TrainingArguments} class in the Hugging Face Transformers library.
\paragraph{Settings for \S\ref{sec:experiment}:}
For BERT-base-japanese, a batch size of 16 and a learning rate of 3e-05 were used, whereas for BERT-large-japanese, a batch size of 8 and a learning rate of 5e-05 were used. 
\paragraph{Settings for \S\ref{sec:ex2}:}
Table~\ref{tbl:appndix_hypara} shows the hyperparameters for BERTs. Eventually, we built five factor-level classifiers in total.

\section{Finetuning Procedure in \S\ref{sec:ex2}}
\label{sec:appendix_finetuning}
Regarding the BERT-based models, we finetuned them to solve the classification tasks.
\paragraph{Dataset:}
In our human annotation process using the decision tree, it is not obligatory for each instance of annotation to respond to all questions at the factor level. 
For example, if an annotator answers ``True'' for Q2 in Figure~\ref{fig:appendix_decision_tree}, he/she does not need to answer any further factor-level questions.
Moreover, each worker might follow a distinct decision-making path for the same annotation instance.
Therefore, the number of answers $n$ to each factor-level question may vary ($0 \leq n \leq 5$).

Consequently, for each set of responses gathered for each factor, we treated it as a binary classification instance only under the condition that the majority label within these $n$ responses can be determined; this approach resulted in a reduction of the dataset size as depict in Table~\ref{table:factor_distribution}.
Eventually, a very small number of samples remained for Factor 3, and therefore, Factor 3 was excluded from the experiment.

\paragraph{Evaluation:}
In order to align the evaluation of BERT with that of human workers, which was measured by the average macro-$F_1$ score of five workers (\S\ref{subsec:statistics}), we computed the average $F_1$ score of five fine-tuned models.

\end{document}